\documentclass[a4paper,article,10pt]{memoir}
\usepackage[english]{babel}
\usepackage{CCLAuthor}
\usepackage[reqno]{amsmath}% AMS-LaTeX-first 
\usepackage{amssymb}
\usepackage{amsthm}  
\theoremstyle{plain}

\theoremstyle{definition}

\numberwithin{theorem}{chapter}% AMS-LaTeX-last
\usepackage[round,authoryear]{natbib} % bibliography
\usepackage[]{graphicx}               % graphics

\newcommand{\MVC}{\operatorname{MVC}}
\newcommand{\push}{\operatorname{push}}
\newcommand{\pull}{\operatorname{pull}}
\newcommand{\load}{\operatorname{load}}
\newcommand{\cem}{\operatorname{cem}}
\newcommand{\shoulder}{\operatorname{shoulder}}
\newcommand{\elbow}{\operatorname{elbow}}
\newcommand{\F}{\operatorname{F}}
\newcommand{\equals}{\stackrel{\mathrm{def}}{=}}
\newcommand{\joint}{\operatorname{joint}}

\begin{document}
%
%----------------------------------------------
%
%    declare the TITLE of your contribution
%    (see Contribs.tex)
%
\title{A new approach to muscle fatigue evaluation for Push/Pull task}
%    
%    AUTHOR(s) and AFFILIATION(s)
%    (see Contribs.tex for a nontrivial example with
%    two authors and three affiliations)
%
\author{%
    Ruina Ma \textsuperscript{\ddag}, Damien Chablat \textsuperscript{\ddag} and Fouad Bennis \textsuperscript{\ddag}
    % end authors
    \\ \smallskip\small% some space
    % begin affiliations
    \textsuperscript{\ddag}
    44321 France (corresponding author: +33240376963; fax: +33240376930; e-mail: Ruina.Ma@irccyn.ec-nantes.fr).
    }% end affiliations
    \maketitle
%
%----------------------------------------------
%
%   ABSTRACT: comment the lines between ABS-first and ABS-last 
%   if your contribution has no abstract
%    
% ABS-first
    \begin{abstract}
    Pushing/Pulling tasks is an important part of work in many industries. Usually, most researchers study the Push/Pull tasks by analyzing different posture conditions, force requirements, velocity factors, etc. However few studies have reported the effects of fatigue. Fatigue caused by physical loading is one of the main reasons responsible for MusculoSkeletal Disorders (MSD). In this paper, muscle groups of articulation is considered and from joint level a new approach is proposed for muscle fatigue evaluation in the arms Push/Pull operations. The objective of this work is to predict the muscle fatigue situation in the Push/Pull tasks in order to reduce the probability of MSD problems for workers. A case study is presented to use this new approach for analyzing arm fatigue in Pushing/Pulling.
    \end{abstract}
% ABS-last
%
%----------------------------------------------
%   CCLsectionING commands
%   numbered CCLsection:          \CCLCCLsection{...}
%   unnumbered CCLsection:        \CCLCCLsection*{...}
%   
%   numbered subCCLsection:       \CCLsubCCLsection{...}
%   unnumbered subCCLsection:     \CCLsubCCLsection*{...} 
%   
%   there are only unnumbered 
%   subsubCCLsections             \CCLsubsubCCLsection{...}
%   
%----------------------------------------------  
%   template for figure placement
%   
%\begin{figure}[htbp]
%            \centering
%        \includegraphics{file name}
%        \caption{...}
%	\label{...}
%    \end{figure}
%----------------------------------------------  
%   template for table placement
%   
%    \begin{table}[htbp]
%        \centering
%        \caption{...}
%        \begin{tabular}{|c|c|...}
%            \hline
%            ... & ...  \\ % row 1
%            \hline
%            ... & ...  \\ % row 2
%        \end{tabular}
%        \label{...}
%    \end{table}
%---------------------------------------------- 

%%%%%%%%%%%%%%%%%%%%%%%%%%%%%%%%%%%%%%%%%%%%%%%%%%%%%%%%%%%%%%%%%%First part
\CCLsection{Introduction}

Approximately 20\% of over-exertion injuries have been associated with push and pull acts~\citep{Cincinnati1981}. Nearly 8\% of all back injuries and 9\% of all back strains and sprains are also associated with pushing and pulling~\citep{Klein1984}. Thus, there is a need to understand push and pull activities in industry since many over-exertion and fall injuries appear to be related to such activities. However, unlike other operations, push/pull capabilities have been studied only within a very limited scope. Most studies describe a laboratory experiment designed to mimic a working condition. For example: Mital did the research of push and pull isokinetic strengths from moving speed and arm angle changes~\citep{Mital1995ClinicBiomechanics}. Badi did an experiment of one handed pushing and pulling strength at different handle heights in the vertical direction~\citep{Badi2008}. 

In industry a number of push/pull operations are causing the MSD problems. From the report of Health, Safety and Executive and report of Washington State Department of Labor and Industries, over 50\% of workers in industry have suffered from MSD, especially for manual handling jobs. According to the analysis in Occupational Biomechanics, muscle fatigue is an essential factor in MSD problems~\citep{Chaffin1999}. Several muscle fatigue models have been proposed in the literature. In Wexler \emph{et al.}, a muscle fatigue model is proposed based on $Ca^{2}+$ cross-bridge mechanism and verified the model with simulation experiments~\citep{Wexler1997}. In Liu \emph{et al.}, a dynamic muscle model is proposed based on motor units pattern of muscle from the biophysical point of view~\citep{Liu2002BiophsicalJournal}. It demonstrates the relationship among muscle activation, fatigue and recovery. Another muscle fatigue model is developed by Giat based on force-pH relationship~\citep{Giat1993}. This fatigue model was obtained by curve fitting of the pH level with time $t$ in the course of stimulation and recovery. In Ma \emph{et al.}, a muscle fatigue model is proposed from the macroscopic point of view~\citep{Ma2009JournalofErgo}. External physical factors and personal factors are taken into consideration to construct the model from joint level. The existing muscle fatigue models consider the muscle fatigue problem from different scientific domain perspectives and each with its own advantages and disadvantages.

We note from the above investigation that few people consider the fatigue factor in the push/pull operation. Therefore, it is necessary to combine the fatigue factor with push/pull operation. The objective of this paper is to introduce a new approach for muscle fatigue evaluation in push and pull operation and it is illustrated by using Push/Pull arm motion. Firstly, we extend the existing static muscle fatigue model in dynamic situation. Secondly, we give some hypothesis of arm muscle activities in push/pull operation. Thirdly, a new approach for muscle fatigue evaluation is proposed. Finally, a case-study is discussed to explain the two articulation push/pull operation of the arm fatigue situation in both shoulder and elbow joints.

%%%%%%%%%%%%%%%%%%%%%%%%%%%%%%%%%%%%%%%%%%%%%%%%%%%%%%%%%%%%%%%Second part
\CCLsection{A New Approach for Push/Pull Arm Fatigue Evaluation}

\CCLsubsection{Human muscle fatigue model}
%%%%%%%%%%%%%%%%%%%%%%%%%%%%%%%%%%%%%%%%%%%%%%%%%%%%%%%%%%%%%% New version
Ma proposed a muscle fatigue model from a macroscopic point of view \citep{Ma2009JournalofErgo}. This muscle fatigue model is expressed as follows.
\begin{equation}
		F_{\cem}(t) =  \MVC\ \cdot\ e^{ \int_{0}^{t} -k  \frac{  F_{\load}(u)}{\MVC} du }
	  \label{eq:fatigue3}
\end{equation}
where $\MVC$ is the maximum voluntary contraction, $F_{\cem}$ is the current exertable maximum force, $F_{\load}(u)$ is the external load and $k$ is a constant parameter. Since Ma \emph{et al.} consider only a static situation, $F_{\load}(u)$ is assumed as a constant. This muscle fatigue model was validated in the context of an industrial by an experimental procedure.

In dynamic situation the value of $F_{\load}$ is not a constant. According to robotic dynamic model \citep{Khalil2010}, $F_{\load}$ can be modeled by a variable depending on the angle, the velocity, the acceleration and the duration of activities.
\begin{equation}
	\F_{\load} \equals \F (u, \theta, \dot{\theta}, \ddot{\theta})
\end{equation}
This way, Eq.~\eqref{eq:fatigue3} can be further simplified in the form.
\begin{equation}
	\F_{\cem}(t) = \MVC\ \cdot\ e^{ -\frac{k}{\MVC} \int_{0}^{t}\F(u, \theta, \dot{\theta}, \ddot{\theta}) du} 
	\label{eq:fatigue6}
\end{equation}
Eq.~\eqref{eq:fatigue6} defines the muscle fatigue model in dynamic situation. The model takes consideration of the motion by the variations of the force $\F_{\joint}$ from joint level. This dynamic factor can be computed using Newton-Euler method or Lagrange method.

\CCLsubsection{Push/Pull muscles activity hypothesis}

\begin{figure}[htb]
   \centering
   {\includegraphics[scale=0.28]{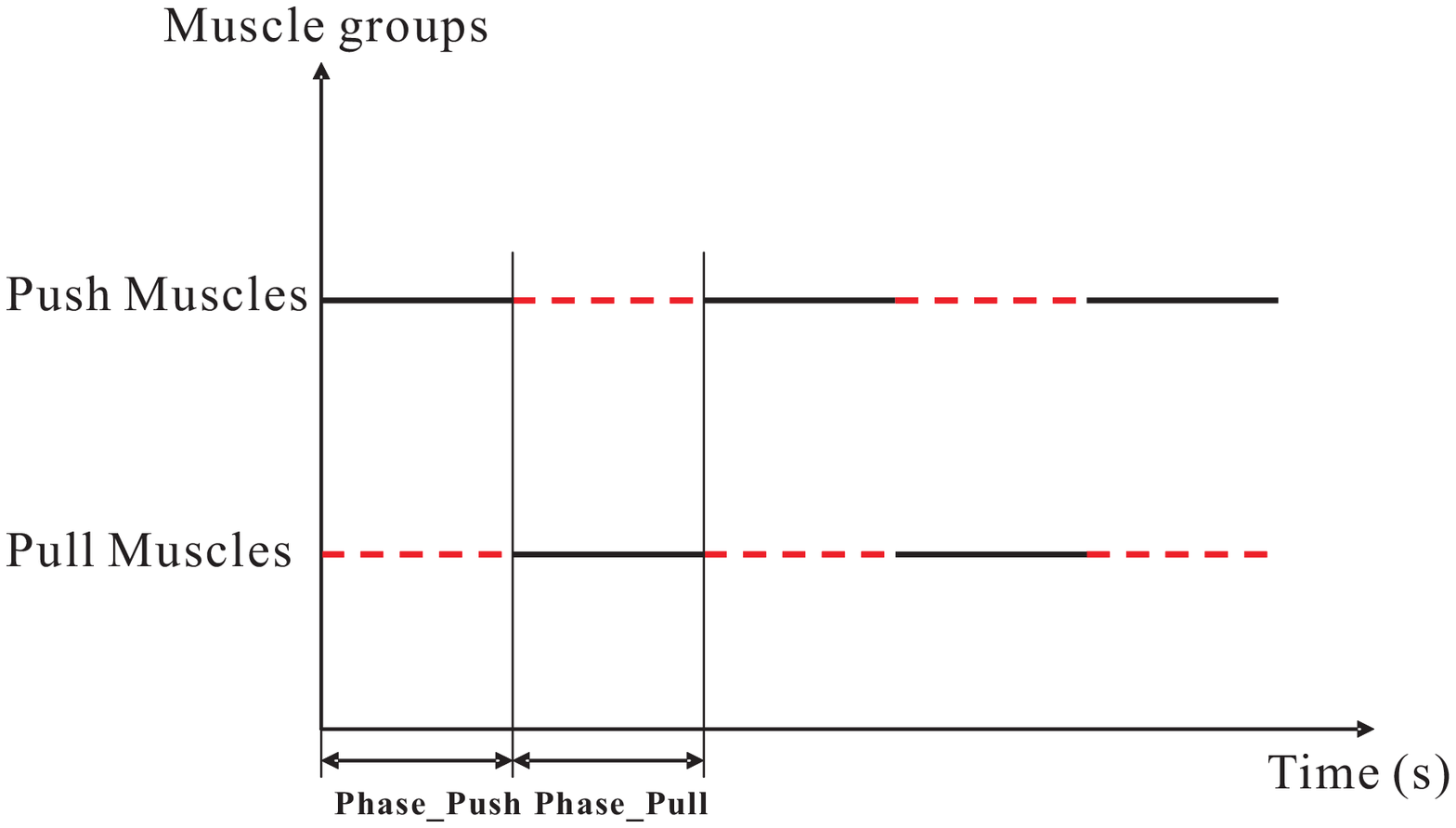}}{\small (a)}
   {\includegraphics[scale=0.28]{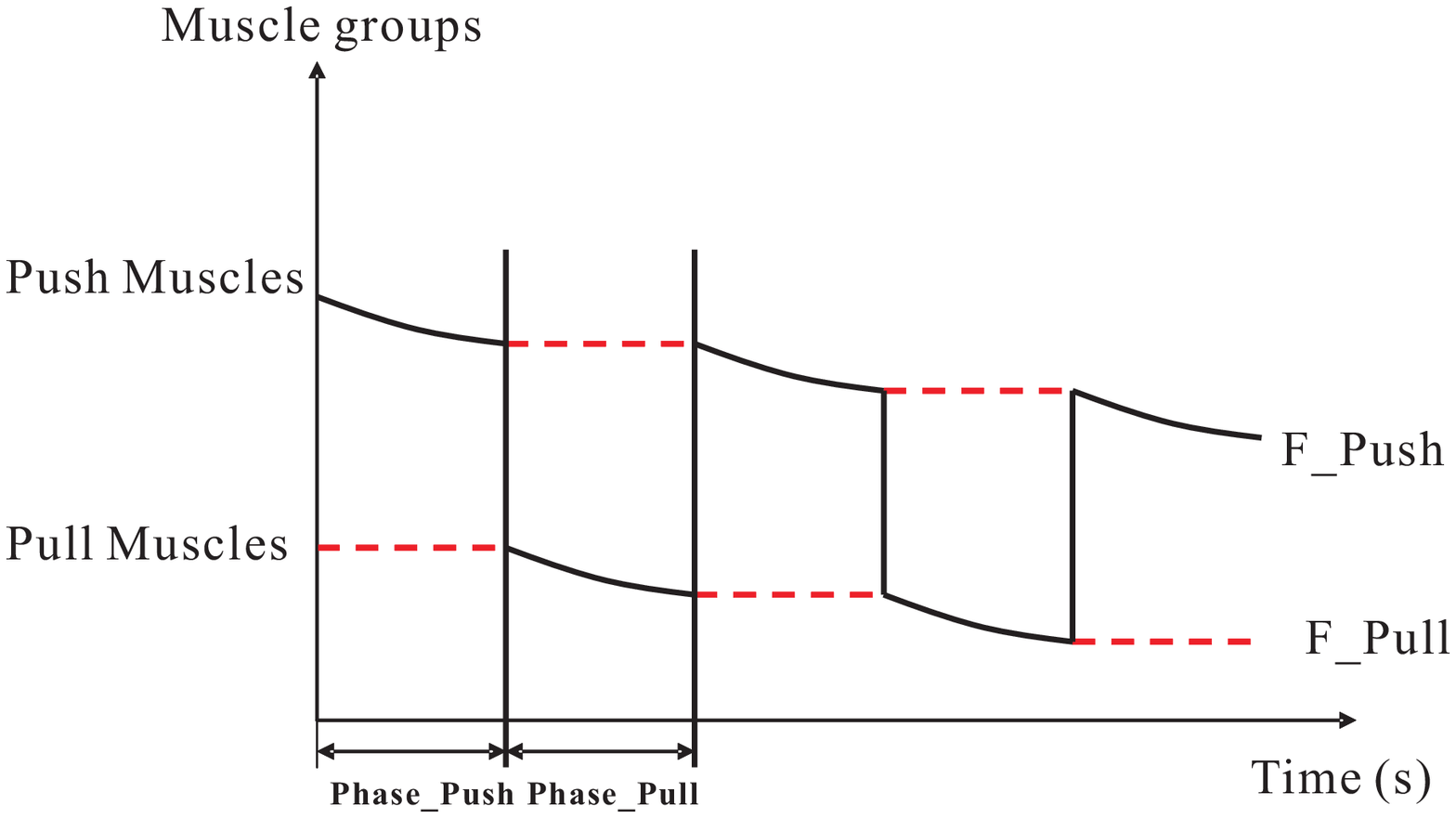}}{\small (b)}
   \caption{\label{fig:Push_Pull_Suppose}(a) Activity of muscle Push/Pull groups (b) Activity of muscle Push/Pull groups with fatigue }
\end{figure}

When a human performs a simple motion, nearly all aspects in the body change. However, from a macroscopic point of view, when a human performs a push/pull operation the related muscle groups of triceps, biceps and deltoid, subscapularis are activated differently. For elbow, when a person does the push operation, the triceps muscle groups are mainly used. Inversely, when a person does the pull operation, the biceps muscle groups are mainly used. For shoulder,  the movement is more complex. Generally when a person does the push operation, we can see that the deltoid related muscle groups are mainly used and inversely in the pull state the subscapu\-la\-ris and other related muscles are mainly used. Based on this, we suppose that every articulation is controlled by two groups muscles: \emph{Push muscles group and Pull muscles group}. Furthermore, we suppose that during the push phase the push muscles group is in an active state and the pull muscles group is in an inactive state. The opposite would occur if the activity is inversed. Here we do not take consideration of the co-contraction factor. We can use Fig.~\ref{fig:Push_Pull_Suppose}(a) to illustrate the activity of push/pull muscles groups. This supposition provides the basis of the new muscle fatigue evaluation approach for push/pull tasks.
\CCLsubsection{Push/Pull muscle fatigue evaluation approach}

From the muscle fatigue model we can see that when a muscle is in an active state, the force generated by muscle decreases with time. In a dynamic push/pull task, the push and pull muscles groups will also experience the fatigue procedure. Based on the muscle fatigue model we introduced before, for a push muscles group, the force exerted trend with time is:%like Eq.~\eqref{eq:push}.
\begin{equation}
	\F_{\push\_\cem}(t) = \MVC_{\push}\ \cdot\ e^{ -\frac{k}{\MVC_{\push}} \int_{0}^{t}\F_{\push}(u, \theta, \dot{\theta}, \ddot{\theta}) du} 
	\label{eq:push}
\end{equation}
Meanwhile for a pull muscles group, the force exerted trend with time is: %like Eq.~\eqref{eq:pull}:
\begin{equation}
	\F_{\pull\_\cem}(t) = \MVC_{\pull}\ \cdot\ e^{ -\frac{k}{\MVC_{\pull}} \int_{0}^{t}\F_{\pull}(u, \theta, \dot{\theta}, \ddot{\theta}) du} 
		\label{eq:pull}
\end{equation}
Push/Pull task is an alternating activity. In this situation, the arm fatigue in push/pull task can be expressed by a piecewise function, as follows: %, like Eq.~\eqref{eq:push_pull}. Figure~\ref{fig:Push_Pull_Suppose}(b) shows the arm fatigue situation in push/pull operation.
\begin{equation}
		F_{\push/\pull}(t) =
			\begin{cases}	
			F_{\push\_\cem}(t), & t \in \operatorname{Phase\_Push}\\
			F_{\pull\_\cem}(t), & t \in \operatorname{Phase\_Pull}
			\end{cases}
	\label{eq:push_pull}
\end{equation}
Figure~\ref{fig:Push_Pull_Suppose}(b) shows the arm fatigue situation in push/pull task.

As for the arm fatigue from a joint-based point of view, we can calculate the force of every articulation (shoulder, elbow) by using a robotic method where every external push and pull force is known. According to the Push/Pull muscle fatigue evaluation approach, when we know the external push/pull force and with other arm biomechanic parameters we can know the arm fatigue trends at both the shoulder and elbow levels.

%%%%%%%%%%%%%%%%%%%%%%%%%%%%%%%%%%%%%%%%%%%%%%%%%%%%%%%%%%%%%%%%%%% Part 3
\CCLsection{Case study: two articulation push/pull operation}

In this case study we discuss the fatigue situation of a two articulation arm in the push/pull operation. From the joint-based view, the arm fatigue can depend on the shoulder fatigue and elbow fatigue. Below are the detailed explanations.

%%%%%%%%%%%%%%%%%%%%%%
\CCLsubsection{Arm model}

\begin{figure}[htb]
   \centering
   {\includegraphics[scale=0.22]{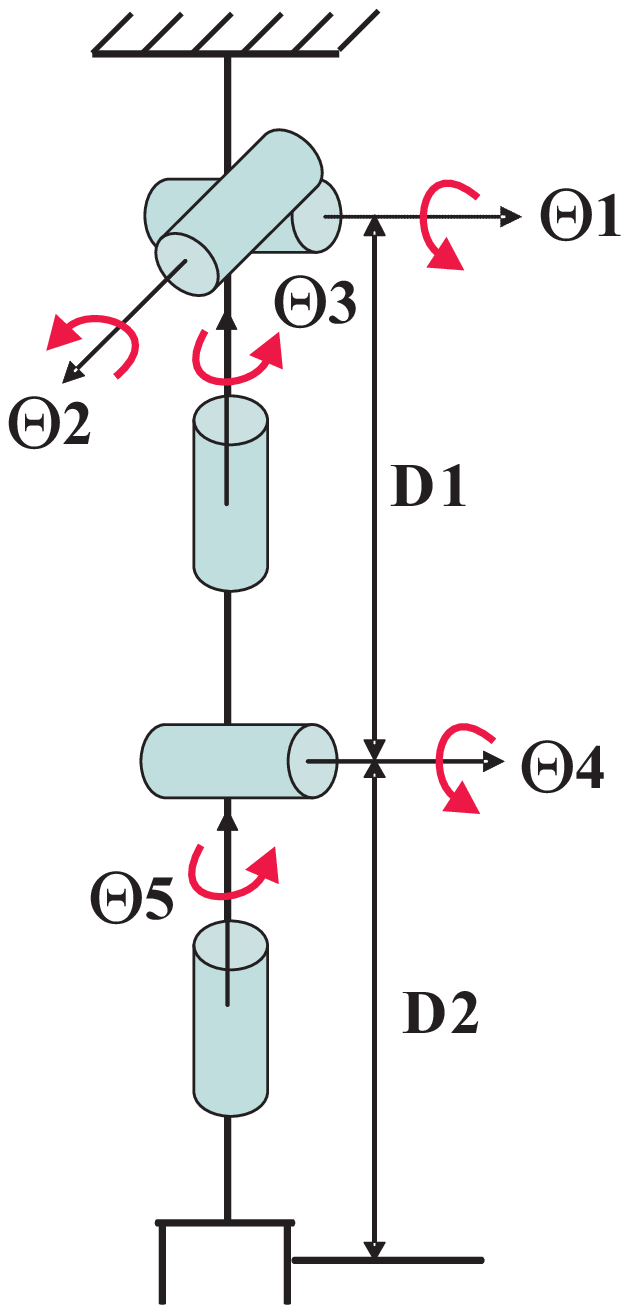}}{\small (a)}
   {\includegraphics[scale=0.30]{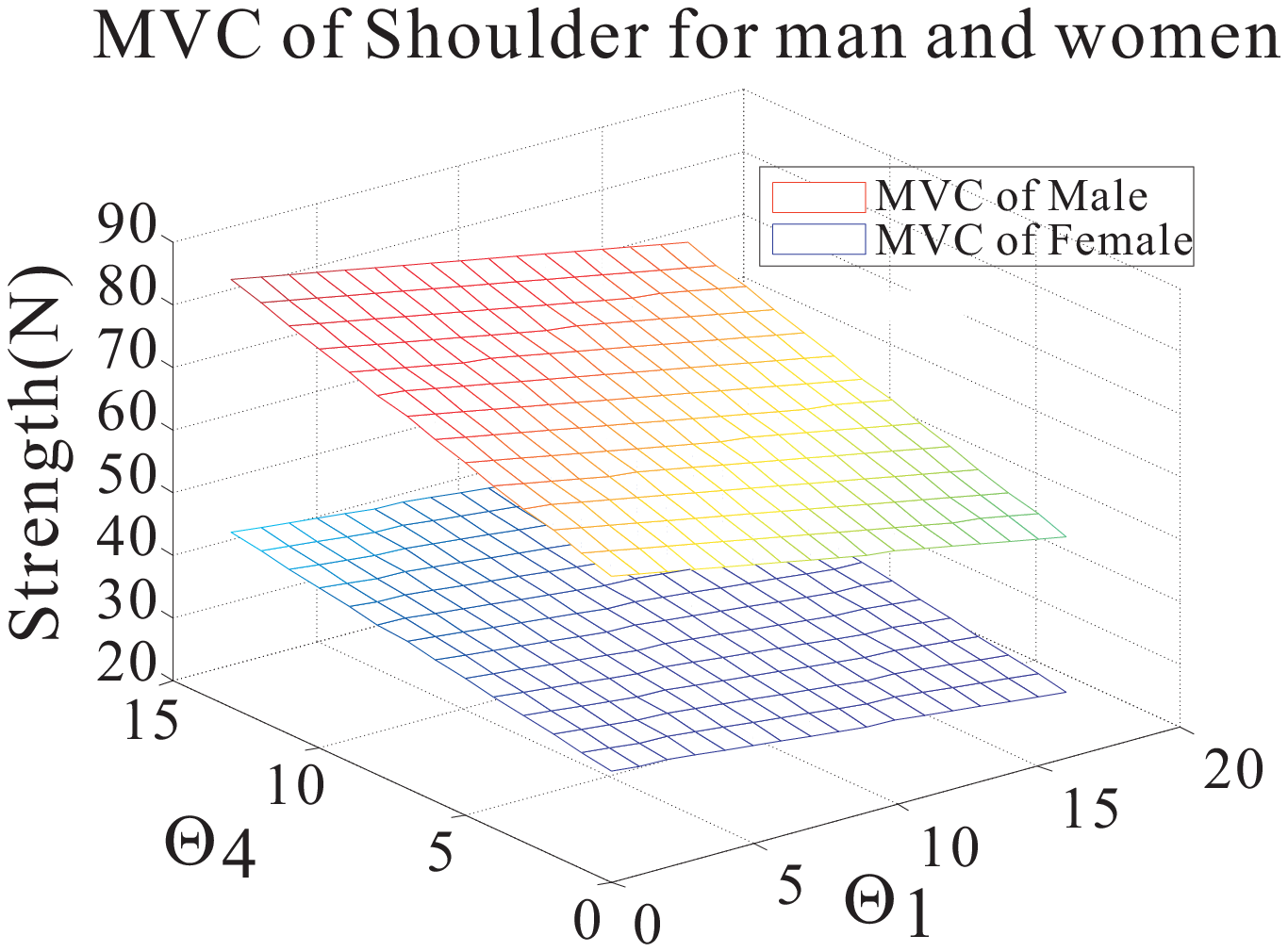}}{\small (b)}
   {\includegraphics[scale=0.30]{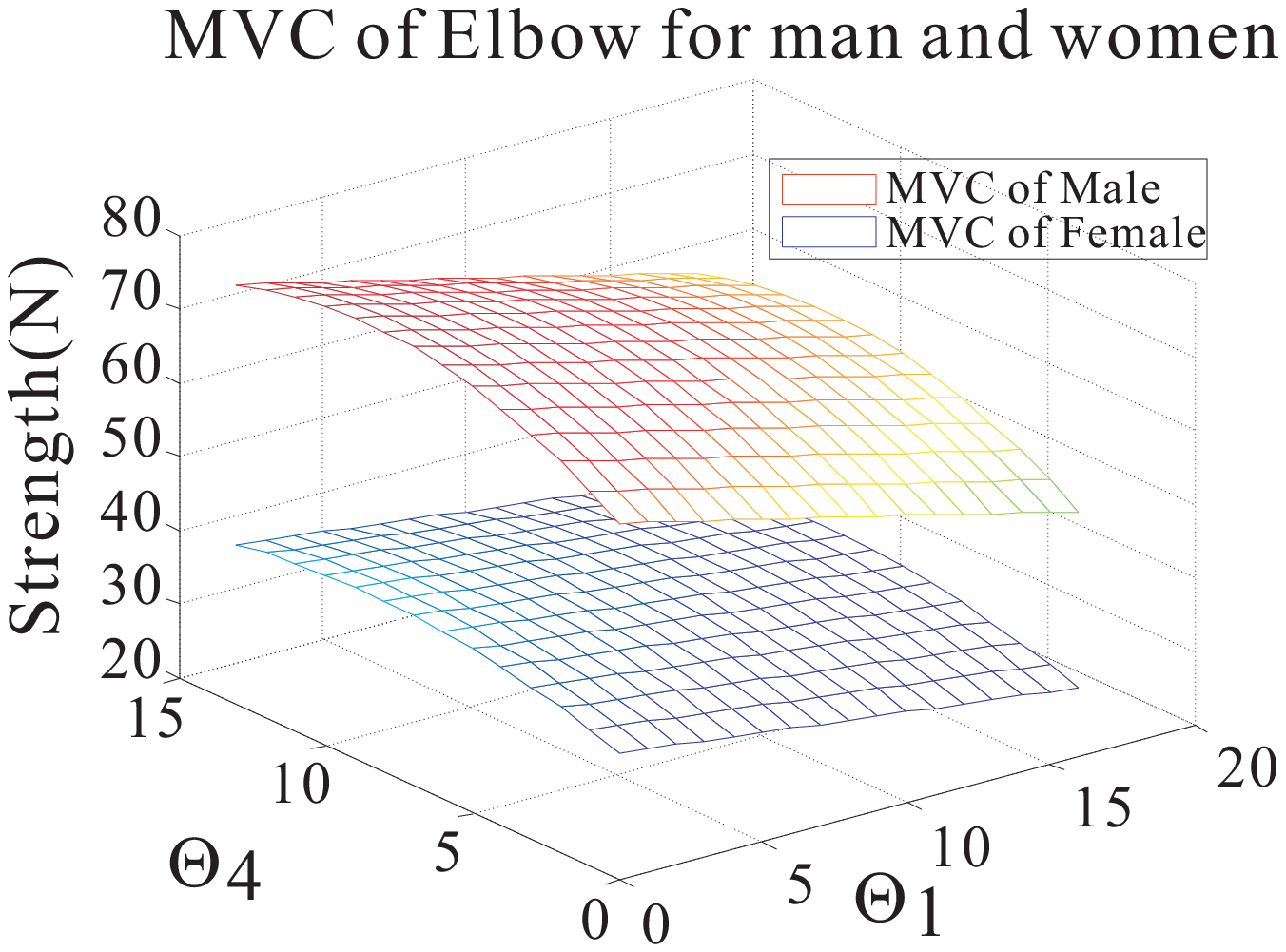}}{\small (c)}
   \caption{\label{fig:strengthModel} (a) Arm geometric model (b) Shoulder MVC of male and female (c) Elbow MVC of male and female~\citep{Chaffin1999} }
\end{figure}

%A geometric model of the arm is first presented to prepare for the later arm fatigue analysis. 
According to the biomechanical structure, we present a geometric model of the arm. We simplify the human arm to 5 revolute joints. The geometric model is shown on Fig.~\ref{fig:strengthModel}(a).

%%%%%%%%%%%%%%%%%%%%%%%%%%%%%%%%%%%%%%%%%%%%%%%%%%%%%%
\CCLsubsection{Shoulder and elbow maximum strength}

In the muscle fatigue equation we should pay attention especially to the $\MVC$ parameter. $\MVC$ means the maximum force that muscles generate. In static situation the $\MVC$ value is a constant, because it is decided by arm posture. Obviously, in a dynamic situation the $\MVC$ value changes with different arm postures. Using the previously researched strength model of Chaffin~\citep{Chaffin1999}, we can get the maximum arm force in different postures. The strength model can be expressed by Eq.~\eqref{eq:strengthModel}, where $\alpha_{e} = \theta_{1}$ and $\alpha_{s} = 180 - \theta_{4}$. $G$ is the parameter for gender adjustment. Fig.~\ref{fig:strengthModel}(b,c) shows the strength of elbow and shoulder in different arm angles both for man and women.
\begin{equation}
	\small
	\begin{array}{ll}
	\textrm{Strength}_{\elbow} 		& = (336.29 + 1.544\alpha_{e} - 0.0085\alpha^{2}_{s})\cdot G \\
	\textrm{Strength}_{\shoulder} & = (227.338 + 0.525\alpha_{e} - 0.296\alpha_{s})\cdot G
	\end{array} 
	\label{eq:strengthModel}
\end{equation}
%
%\input{figures/ArmPosture}

%%%%%%%%%%%%%%%%%%%%%%%%
\CCLsubsection{Fatigue risk force}

\begin{figure}[htb]
   \centering
   {\includegraphics[scale=0.3]{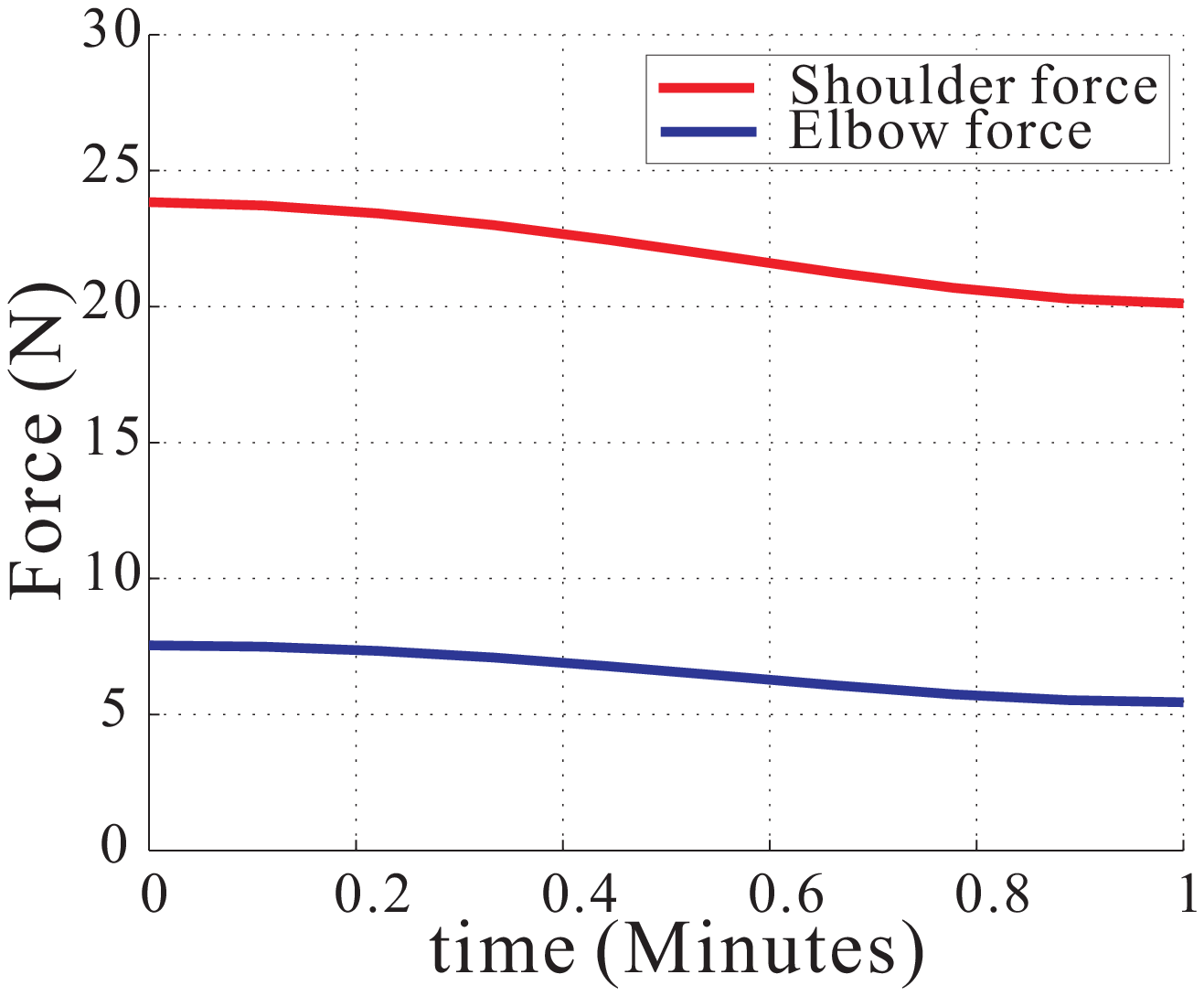}}{\small (a)}
   {\includegraphics[scale=0.3]{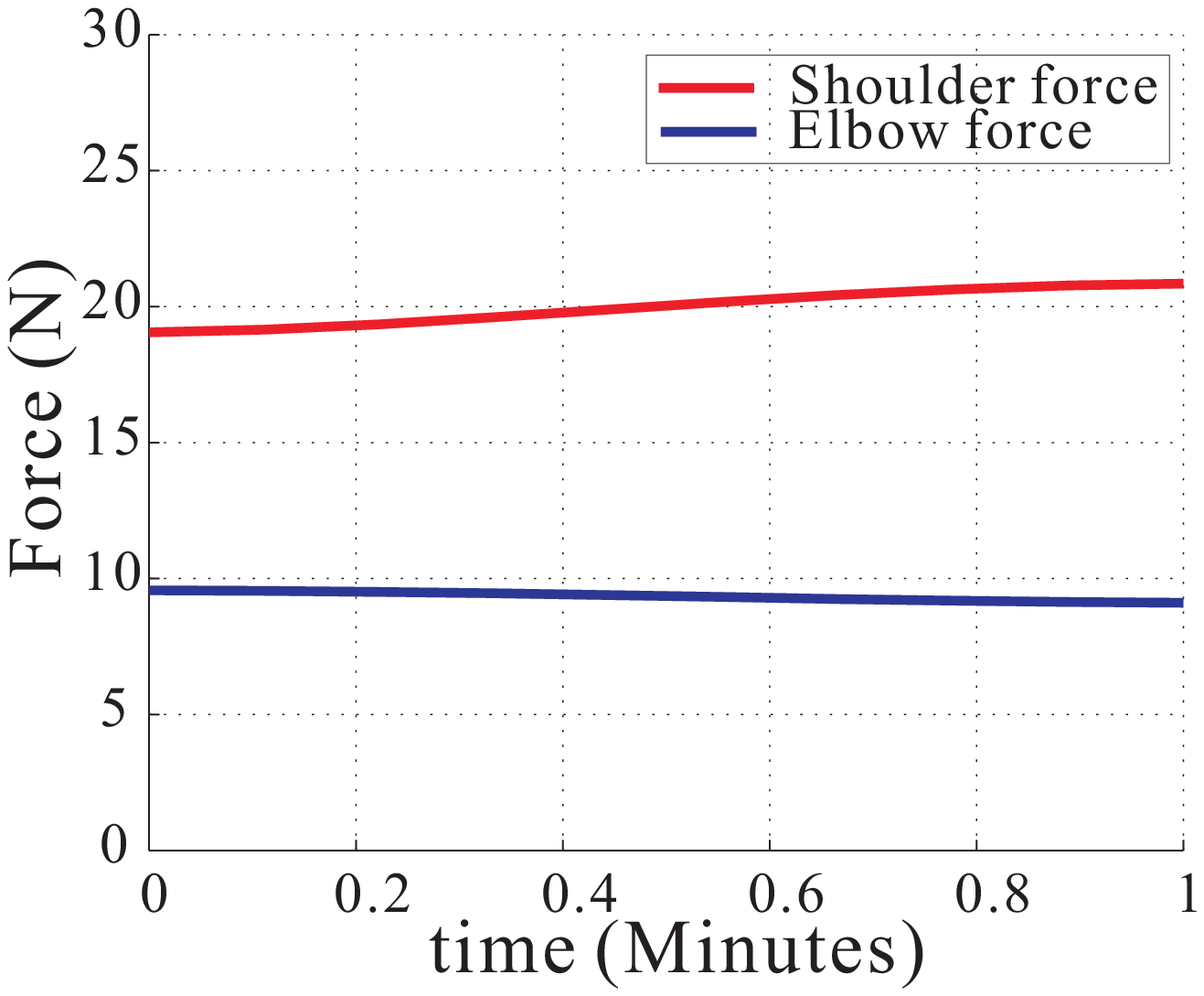}}{\small (b)}
   \caption{\label{fig:fatigue_risk_line}(a) Push Operation (b) Pull Operation }
\end{figure}

When people do some tasks, at what point will they risk fatigue? A subjective measure of fatigue is not appropriate to quantify fatigue. We need an objective parameter to measure the fatigue risk. The Maximum Endurance Time (MET) is an important concept in ergonomics. It describes the duration from the start to the instant at which the strength decreases to below the torque demand resulting from external load. Once the external load exceeds the current force capacity, potential physical risks might occur to the body tissues. So the fatigue risk force, noted $F_{\operatorname{fatigue\_risk}}$ is reached when the external load equals the current force capacity, 
\begin{equation}
F_{\operatorname{fatigue\_risk}} = {F}_{\operatorname{external}} = {F}_{\operatorname{joint}}. 
\end{equation}
For a static posture, $F_{\operatorname{fatigue\_risk}}$ is a constant value, because in the static situation the external force is a constant. Meanwhile, in dynamic situation,  $F_{\operatorname{fatigue\_risk}}$ is non-linear function; it changes according to different postures. According to the arm geometric model, the force of every joint can be represented by $F_{\operatorname{joint}} \in \{ F_{\theta_1},\ldots,F_{\theta_5} \}$.  For example, $F_{\operatorname{fatigue\_risk}}$ of shoulder is $F_{\shoulder} = F_{\theta_{1}}$ and $F_{\operatorname{fatigue\_risk}}$ of elbow is $F_{\elbow} = F_{\theta_{4}}$. When the arm does the horizontal push and pull operation and the hand move alone $x$ axis between (0.3, 0.4) meter, the shoulder angle changes between (-49.3, -42.3) degrees and the elbow angle changes between (124.1, 101.3) degrees. The force changes of shoulder and elbow were shown on Fig.~\ref{fig:fatigue_risk_line}. This also represents the fatigue risk force trends in the push/pull operations. Forces of shoulder and elbow are computed using Newton-Euler method~\citep{Khalil2010}.
%
%%%%%%%%%%%%%%%%%%%%%%%%%
\CCLsubsection{Dynamic push/pull simulation}

\begin{figure}[htb]
   \centering
   {\includegraphics[scale=0.32]{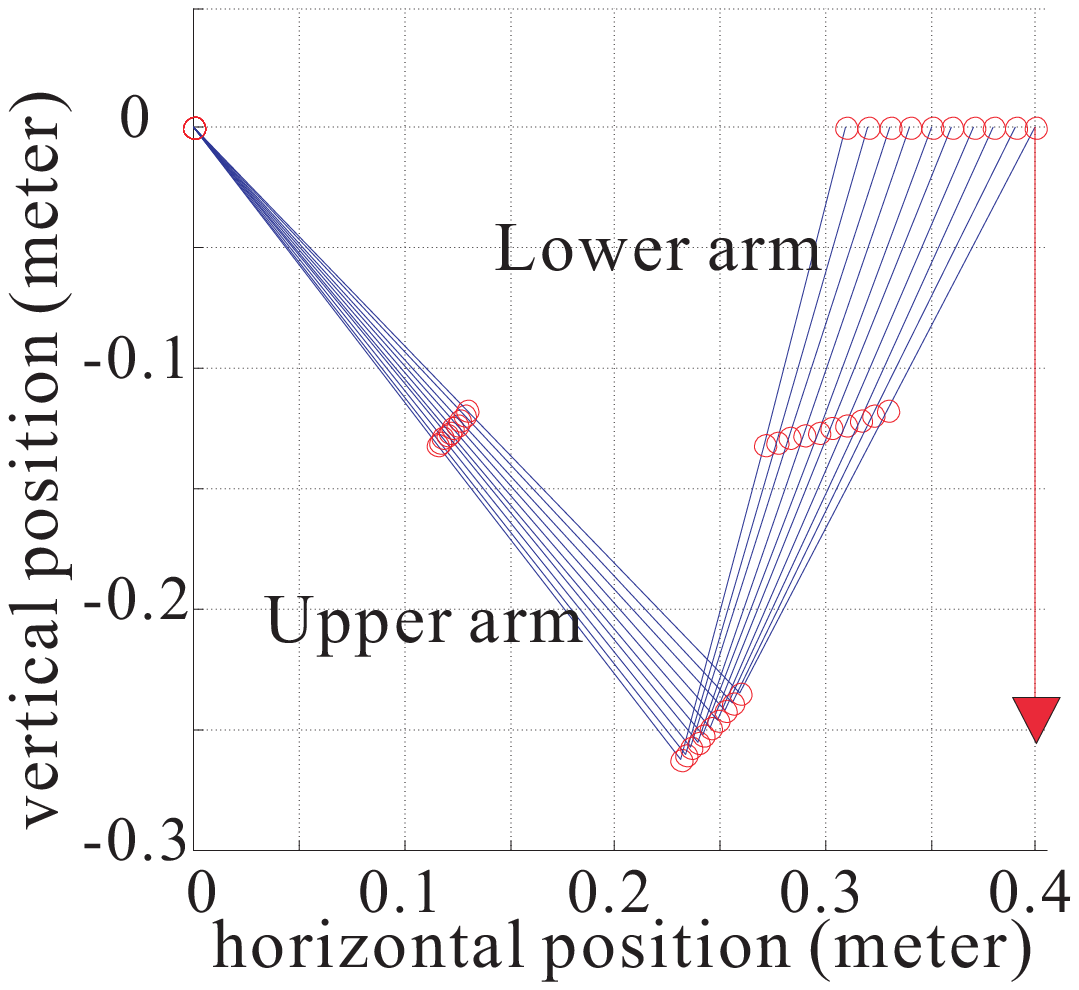}}{\small (a)}
   {\includegraphics[scale=0.31]{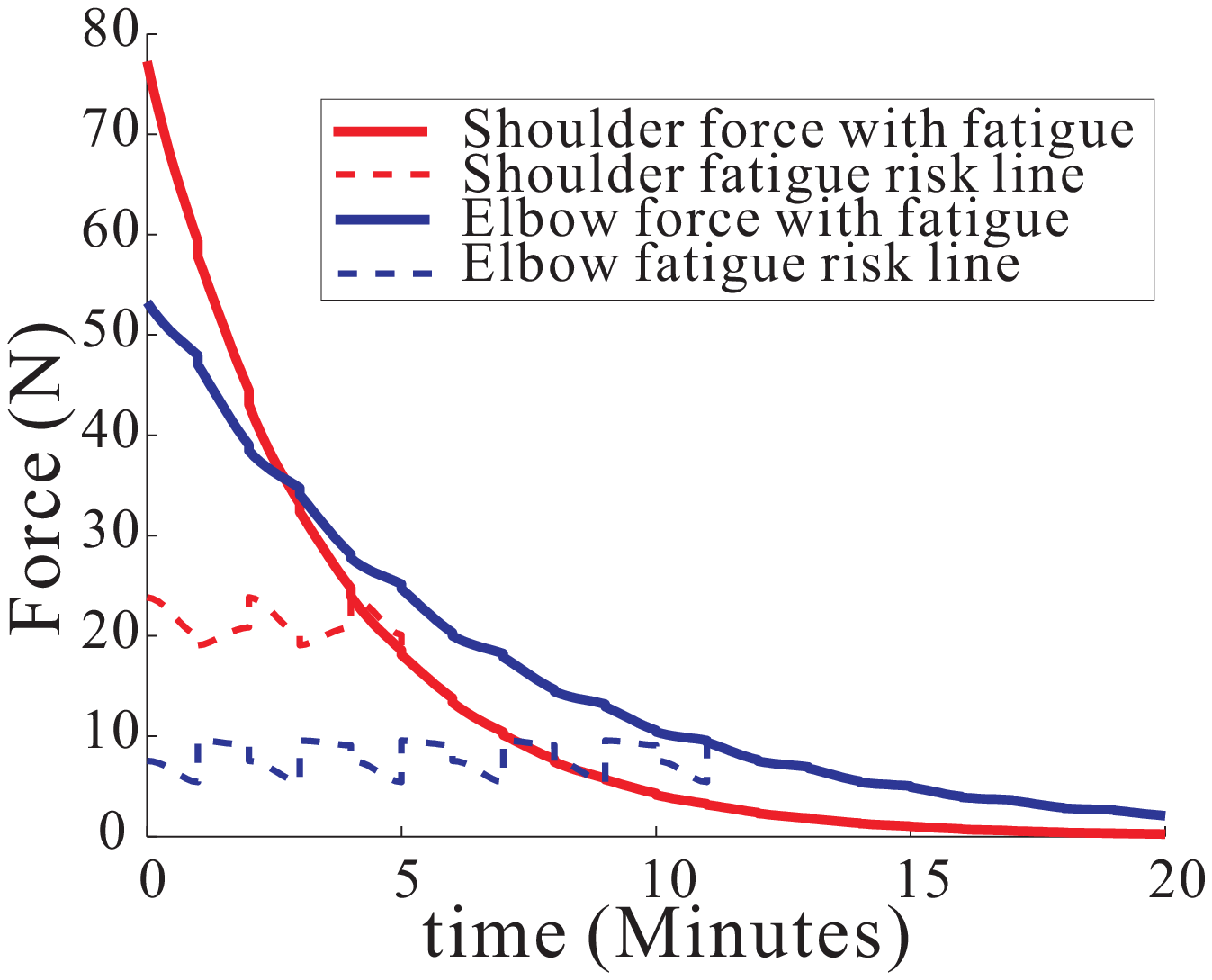}}{\small (b)}
   \caption{\label{fig:Push_Pull_Simulation}(a) Arm Push/Pull Operation (b) Fatigue simulation of arm Push/Pull Operation }
\end{figure}

Suppose the weight of the object is 2Kg, and a person (Sex:~man, Height:~188cm, Weight:~90kg) uses 10N to push and 10N to pull this object. In reality the push/pull task time may not be exactly the same, here we simplify the situation and consider that the two operations use the same time ($T_{\push} = T_{\pull}$ = 1 minute). Fatigue rate parameter $k$ depends on the subject and experiment measures are needed to determine it. Here for this specific case we choose $k = 1$ for both push and pull operations. In different postures, according to the estimated arm's strength we know the $\MVC$ of the shoulder and elbow at every moment; according to the push force and pull force we can calculate the force of elbow and shoulder at every moment. Furthermore we can know the fatigue risk line of the shoulder and elbow. Based on these initial conditions, and using our approach, we can determine the fatigue trends in shoulder and elbow separately. Fig.~\ref{fig:Push_Pull_Simulation} shows the simulation of arm fatigue in both shoulder and elbow level.

%%%%%%%%%%%%%%%%%%%%%%%%%%%%%%%%%%%%%%%%%%%
\CCLsubsection{Discussion}
From the simulation result we can see that in push/pull task, the shoulder exerted force reaches $F_{\operatorname{fatigue\_risk}}$ in about 5 minutes and the elbow exerted force reaches $F_{\operatorname{fatigue\_risk}}$ in about 11 minutes. This explains why many workers have problems with shoulder rather than elbow from pushing/pulling. These results support the idea that when  workers do this kind of work, it is better to have a rest or to change the working posture every 5 minutes to avoid fatigue. The simulation here is a general case; the specific prediction is decided by the individual person, the working posture, the weight of the object and the single push/pull operation time. In future work we will use experimental data to verify this fatigue evaluation approach. The experiment would also consider more muscles and joints, female subjects and changing external loads.

%%%%%%%%%%%%%%%%%%%%%%%%%%%%%%%%%%%%%%%%%%%%%%%%%%%%%Part 4
\CCLsection{Conclusions}
In this paper, we took into consideration the muscle groups of articulation in push and pull operations and from joint level proposed a new approach to muscle fatigue evaluation for push/pull tasks. From simulation results we can see that in a push/pull process, the arm fatigue trends appear in both shoulder and elbow level and it is decided by the parameters $MVC$, $F_{\load}$, $k$, and working postures. The case study shows that our new fatigue evaluation approach has a potential to provide information to help people prevent fatigue risk and estimate the safe working periods for pushing/pulling. Our work enlarges the vision of push/pull operation research. The final goal of our work is to use this model in ergonomic evaluation procedures, to enhance work efficiency and reduce MSD risks.

%%%%%%%%%%%%%%%%%%%%%%%%%%%%%%%%%%%%%%%%%%%%%%%%%%
%                                                %
%          BIBLIOGRAPHY with or without BibTeX   %
%                                                %
%%%%%%%%%%%%%%%%%%%%%%%%%%%%%%%%%%%%%%%%%%%%%%%%%%
%      
%   BIBLIOGRAPHY without BibTeX
%   Remove the %% from the lines starting with %%
%   if you want to test it
%   
%%\cite{CG91} investigated\ldots % a citation inside your text
%
% near the end of your document you must insert the bibliographic
% items inside a "thebibliography" environment
% 
%%\par\begin{thebibliography}{0}  % environment start
%%    \bibitem[Charniak and Goldman(1991)]{CG91}
%%             E.~Charniak and R.~Goldman, 
%%      \textit{A probabilistic model of plan recognition},
%%      In {\emph Proceedings of the Ninth Conference on Artificial
%%      Intelligence}, 
%%      pages 160--165, 1991.
% other bibitems below
%   \bibitem[...]{...}
% A. U. Thor, \textit{Knuth's TeX and METAFONT},...
%   
%   \bibitem[...]{...}
%%\end{thebibliography}       % environment end
%
%----------------------------------------------  
%  
%   BIBLIOGRAPHY with BibTeX
%   Remove the %% from the lines starting with %%
%   to test it
% 
%%\cite{CG91} investigated\ldots % a citation inside your text
%
%   near the end of your document you must declare the BibTeX
%   database containing the items you cited (Contribs)
%   as well as the bibliographic style {plainnat}
% 
\bibliographystyle{plainnat}
\bibliography{bibilography}   

\end{document}